\documentclass[final]{ecai}
\usepackage[latin9]{inputenc}
\usepackage{xcolor}
\usepackage{array}
\usepackage{float}
\usepackage{booktabs}
\usepackage{textcomp}
\usepackage{multirow}
\usepackage{amsmath}
\usepackage{amsthm}
\usepackage{amssymb}
\usepackage{graphicx}
\PassOptionsToPackage{normalem}{ulem}
\usepackage{ulem}

\makeatletter

\newcommand{\lyxmathsym}[1]{\ifmmode\begingroup\def\b@ld{bold}
  \text{\ifx\math@version\b@ld\bfseries\fi#1}\endgroup\else#1\fi}

\DeclareTextSymbolDefault{\textquotedbl}{T1}
\providecommand{\tabularnewline}{\\}
\floatstyle{ruled}
\newfloat{algorithm}{tbp}{loa}
\providecommand{\algorithmname}{Algorithm}
\floatname{algorithm}{\protect\algorithmname}

\usepackage{latexsym}
\usepackage{amsthm}
\usepackage{enumitem}
\usepackage{color}

\usepackage[linesnumbered,algo2e]{algorithm2e}
\usepackage{algorithm}
\usepackage{algorithmic}
\newcommand{\hrulealg}[0]{\vspace{1mm} \hrule \vspace{1mm}}

\newcommand{\BibTeX}{B\kern-.05em{\sc i\kern-.025em b}\kern-.08em\TeX}

\title{Planner-Refiner: Dynamic Space-Time Refinement for Vision-Language Alignment in Videos}

\makeatother

\begin{document}
\begin{frontmatter}

\global\long\def\ModelName{\text{Planner-Refiner}}%

\global\long\def\DataName{\text{MeViS-X}}%

 \author{\fnms{Tuyen}~\snm{Tran}\orcid{0009-0002-8161-7637}\thanks{Corresponding Author. Email: t.tran@deakin.edu.au}} 
\author{\fnms{Thao}~\snm{Minh Le}\orcid{0000-0002-8089-9962}} 
\author{\fnms{Quang-Hung}~\snm{Le}\orcid{0000-0003-4727-6859}} 
\author{\fnms{Truyen}~\snm{Tran}\orcid{0000-0001-6531-8907}}

\address{Applied Artificial Intelligence Institute, Deakin University, Australia}
\begin{abstract}
Vision-language alignment in video must address the complexity of
language, evolving interacting entities, their action chains, and
semantic gaps between language and vision. This work introduces $\ModelName,$
a framework to overcome these challenges. $\ModelName$ bridges the
semantic gap by iteratively refining visual elements' space-time representation,
guided by language until semantic gaps are minimal. A Planner module
schedules language guidance by decomposing complex linguistic prompts
into short sentence chains. The Refiner processes each short sentence---a
noun-phrase and verb-phrase pair---to direct visual tokens' self-attention
across space then time, achieving efficient single-step refinement.
A recurrent system chains these steps, maintaining refined visual
token representations. The final representation feeds into task-specific
heads for alignment generation. We demonstrate $\ModelName$'s effectiveness
on two video-language alignment tasks: Referring Video Object Segmentation
and Temporal Grounding with varying language complexity. We further
introduce a new $\DataName$ benchmark to assess models' capability
with long queries. Superior performance versus state-of-the-art methods
on these benchmarks shows the approach's potential, especially for
complex prompts. Our code is available at: https://github.com/tranxuantuyen/Planner-Refiner

\end{abstract}
\end{frontmatter}

\section{Introduction}

\begin{figure}[t]
\centering{}\includegraphics[width=0.99\columnwidth]{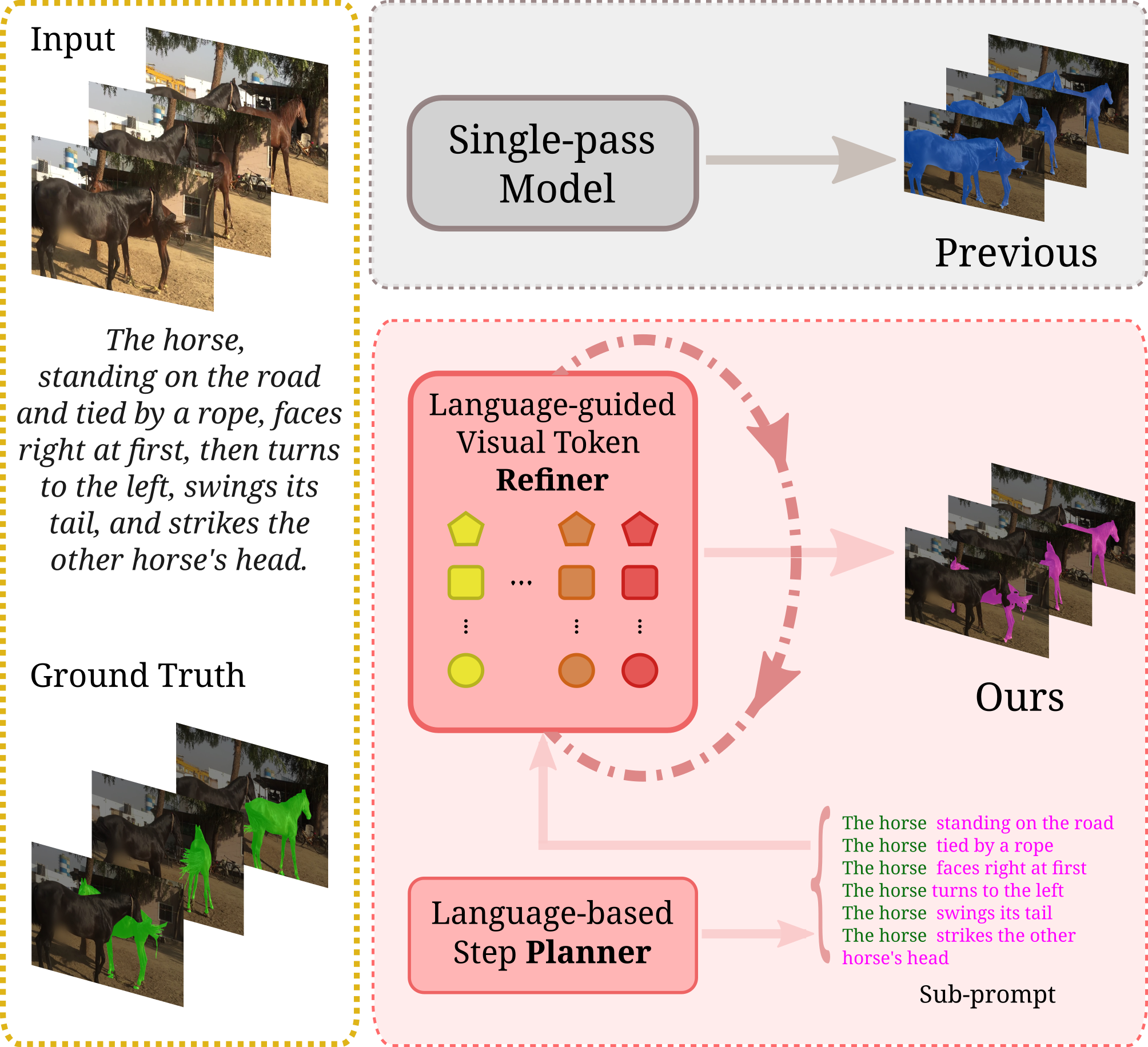}\caption{Our $\protect\ModelName$ (bottom-right), compared with the current
single-pass approach (top-right). Ours is a recurrent system that
iteratively refines the visual representation across space-time conditioned
on the language. The linguistic input of each step is planned by a
Recurrent Step Planner, which decomposes the input prompt into a chain
of short sentences. This mechanism enables more fine-grained refinement
of visual tokens, resulting in improved accuracy in predictions. \label{fig:teaser}}
\end{figure}

Video-language alignment has emerged as a critical challenge in multimodal
understanding, particularly in bridging the semantic gaps between
low-level visual content and linguistic concepts \cite{rizve2024vidla,singh2022flava,tran20242nd,ding2024lsvos}.
While recent large video-language models have demonstrated impressive
capabilities, their effectiveness remains largely limited to matching
simple textual concepts to visual regions \cite{NEURIPS2023_6dcf277e,pmlr-v202-li23q}.
This falls short to meet practical requirements that demand a sophisticated
understanding of complex concepts featuring multiple components within
a single query. Consider a video surveillance scenario \cite{surveillance}
where machines are required to process a complex query like\emph{``Find
in video a person wearing a red jacket enters the room, picks up a
black bag, and exits through the left door''.} Such queries require
simultaneous comprehension of a series of interrelated components
including cross-modality subject and attribute identification, modeling
object dynamics in space-time and understanding cross-object dependencies.

The challenge lies in developing computer systems that are capable
of modeling the interactions of multiple visual entities in space
and time in reflection the interpretation of a given query \cite{Xue2022CLIPViPAP,munasinghe2024videoglamm,tran2024unified}.
This is not a trivial task especially in complex scenes where the
involved entities perform intricate sequences of actions and interactions.
Fig.~\ref{fig:teaser} provides such an example. Here, we consider
the task of identifying a video segment featuring a horse with a sequence
of actions, including \emph{faces right}, \emph{turns to the left},
\emph{swings its tail} and \emph{strikes the other horse's head}.
Current state-of-the-art approaches, such as DsHmp \cite{DsHmp},
can successfully recognize the presence of objects (e.g., horses),
but struggle with comprehending complex action sequences associated
with specific instances.  

There are two fundamental limitations in the current state-of-the-art
approaches. First, base vision-language systems are typically pretrained
on datasets that focus on isolated, single-concept scenarios \cite{pmlr-v202-li23q,radford2021learning,smaira2020short},
thus incapable of reasoning about compositional scenarios involving
complex relationships between elements. Second, current methods often
overly focus on visual modeling while employing oversimplified approaches
to handle complex queries. Works such as \cite{heo2022vita,MeViS,miao2024htr,cheng2021mask2former}
compress entire input sequences into single embeddings, neglecting
crucial linguistic structural information about objects interactions
that goes beyond the ordinary autoregressive modeling of sequence
of words. While DsHmp \cite{DsHmp} partially addresses this limitation
by breaking a textual query into static and motion cues, its rigid
two-component architecture proves inadequate for compositional sentences
with varying complexity levels.

To address these limitations, we introduce $\ModelName$, a novel
framework that minimizes visual-linguistic semantic gaps through iterative
refinement of visual representations in space-time guided by language.
At its core, $\ModelName$ employs two key components: a language-based
step \textbf{Planner} module that decomposes a complex textual query,
also referred to as the referring prompt, into atomic action sequences,
and a following\textbf{ Refiner} recurrent system that progressively
refines visual representations as the action sequences revealed. More
concretely, the Planner dynamically creates a variable-length sequence
of sub-prompts, each comprising a noun phrase (object) and a verb
phrase (action), with the sequence's length determined by the syntactic
structure of the input query. These sub-prompts are ordered in temporal
dimension, reflecting the specific action chain performed by the target
instance. They then serve as guide to the \textbf{Refiner} to iteratively
refine spatial and temporal visual tokens. The refinements of these
visual tokens are performed separately following the English grammar
structure where noun typically comes before verb in a simple sentence.
This architecture enables progressive improvement in understanding
both objects and their actions over time, facilitating more accurate
vision-language alignments. 

We demonstrate the effectiveness of our approach through comprehensive
experiments on two video-language alignment tasks: Referring Video
Object Segmentation (RVOS) and Video Temporal Grounding (VTG). Our
results show that $\ModelName$ achieves a 20\% relative improvement
in segmentation accuracy over the baseline on the RVOS task. Additionally,
we introduce a novel $\DataName$ benchmark, specifically designed
to evaluate model performance on complex, compositional descriptions.
$\ModelName$ demonstrates significant improvements over state-of-the-art
methods on this benchmark, underscoring its effectiveness in handling
complex video-language tasks, particularly those involving dynamic,
temporally extended descriptions.

In summary, our contributions include:
\begin{enumerate}
\item A novel two-component system for vision-language alignment in videos,
consisting of a Language-based Step Planner that dynamically determines
a sequence of reasoning steps based on query complexity, followed
by a Language-guided Visual Token Refiner for language-guided refinement
of visual tokens.
\item The $\DataName$ benchmark for evaluating reasoning capability with
complex sentences in the RVOS task. 
\item Extensive experiments and analysis across video-language alignment
tasks, providing insights into the effectiveness of the proposed approach
towards solving such a challenging task.
\end{enumerate}

\section{Related Work}

Recent large-scale vision-language models \cite{pmlr-v202-li23q,munasinghe2024videoglamm,singh2022flava}
excel at aligning simple textual descriptions with visual data. Yet,
these systems struggle with complex scenarios due to their pretraining
on single-entity events \cite{aflalo2024fivl}. Early works \cite{wang2019asymmetric,huang2020referring}
on video-language alignments extended image-text grounding models
using post-processing techniques to ensure temporal coherence. However,
processing videos in a frame by frame basis neglects inherent motion
information, which is crucial for understanding the dynamics of video
content. Recent approaches \cite{wu2022language,botach2022end,MeViS}
explored end-to-end training for this task by leveraging DETR \cite{carion2020end}
to extract visual tokens, bypassing the need for separate detection
and tracking processes. While having potential, these methods lack
efficient mechanisms to handle complex, compositional input sequences.
The entire sentence is often encoded into single vector embedding
regardless of the level of complexity, neglecting the rich structure
of human language. The current SOTA method, DsHmp \cite{DsHmp} took
a step further by decomposing input sentences into static and motion
cues. However, DsHmp hypothesized a fixed two-component architecture
for all input sequences regardless of their levels of complexity.
In this work, we address these limitations through dynamic decomposition
of complex queries into a set of sub-prompts, each describing an event
of a target instance performing an atomic action found in video data. 

\begin{figure*}[t]
\centering{}\includegraphics[width=0.9\textwidth]{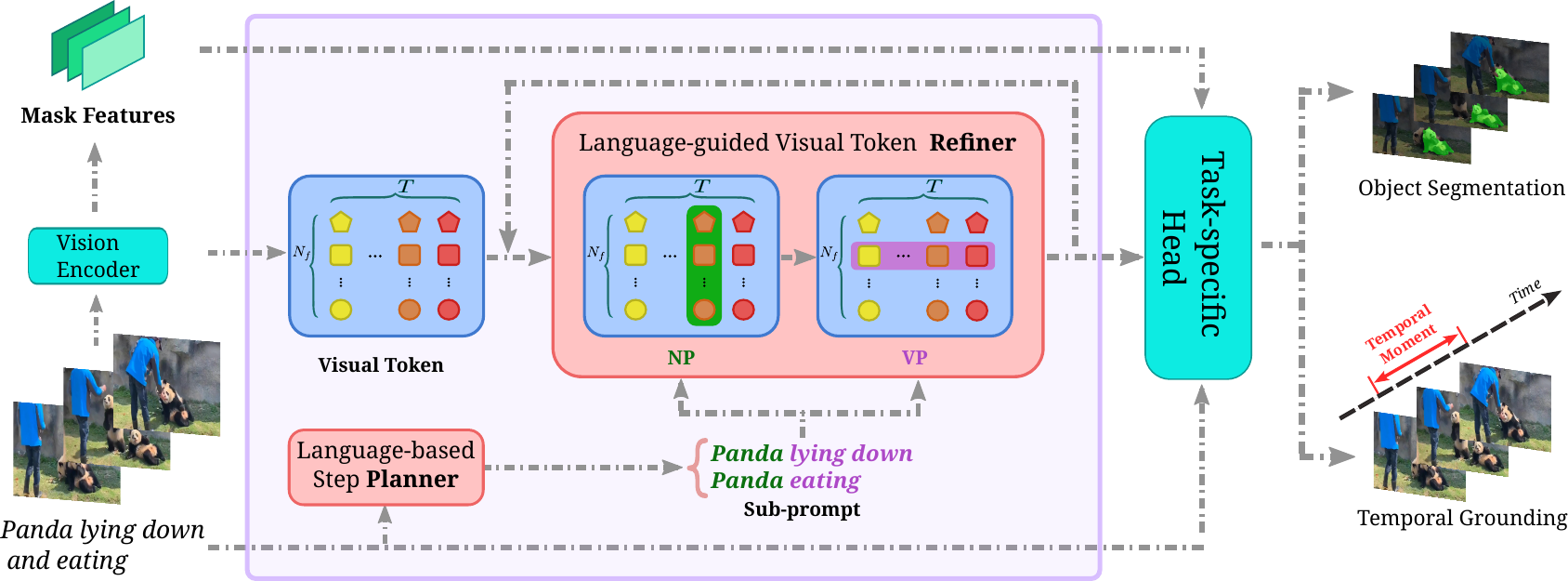}\caption{\textbf{Method overview. }Our framework is composed of three main
stages: (1) The\emph{ Language-based Step }\textbf{\emph{Planner}}
decomposes an input query into a set $P$ sub-prompts, each containing
a noun phrase (\textbf{\textcolor{olive}{NP}}) and a verb phrase (\textbf{\textcolor{magenta}{VP}}).
(2) The \emph{Language-guided Visual Token }\textbf{\emph{Refiner}}\emph{
}progressively refines $N_{f}\times T$ visual tokens over $P$ steps
as guided by the sub-prompts. At each recurrent step $p$, refinement
occurs through specialized spatial attention guided by NP followed
by a temporal attention guided by VP. (3) The Task-specific heads
adapt the refined tokens for downstream tasks - generating segmentation
masks for RVOS or temporal prediction for VTG.\label{fig:method}}
\end{figure*}

In addition to architectural design challenges, the absence of comprehensive
datasets specifically designed for complex query understanding further
hinders research in this area. To bridge this gap, W. Ji et al. \cite{ji2024toward}
introduced the RIS-CQ dataset, specifically designed for the Referring
Image Segmentation task \cite{cheng2023wicowinwincooperationbottomup,ijcai2023p144}
with complex language queries. When it comes to videos, most widely
recognized datasets \cite{seo2020urvos,Gavrilyuk_2018_CVPR} contain
only simple and short descriptions that do not require much of motion
understanding. These limited descriptions enable AI models to achieve
high performance by relying solely on static visual saliency \cite{MeViS}.
While MeViS \cite{MeViS} offers motion-based descriptions for the
task, these descriptions are still short and simple, which typically
contain only one or two actions. Our $\DataName$ benchmark extends
the MeViS dataset with more complex, highly compositional queries.
See Supplementary Material for a detailed literature review on complex
query understanding and the two tasks: Referring Video Object Segmentation
and Video Temporal Grounding.

\section{Method\label{sec:Method}}

We now detail $\ModelName$, a recurrent system performing multi-step
reasoning based on linguistic structure to refine visual token representations
for video-language alignment tasks. Fig.~\ref{fig:method} provides
an overview of our architecture. Following \cite{MeViS,cheng2021mask2former,DsHmp},
we extract spatio-temporal features from videos using a pre-trained
vision encoder \cite{cheng2021mask2former,tran2015learning} (Section
\ref{subsec:Preliminaries}). Our \emph{Language-guided Visual Token}
\textbf{Refiner} iteratively refines these features across multi-stages,
guided by the language components dynamically generated by the \emph{Language-based
Step} \textbf{Planner}, enabling precise text-visual alignment through
rigorous linguistic comprehension (detailed in Sections \ref{subsec:planner}
and \ref{subsec:refiner}). The system is task-agnostic, adaptable
to various video-language alignment tasks via task-specific heads.
We demonstrate this through two applications: Referring Video Object
Segmentation and Video Temporal Grounding, both requiring complex
sentence interpretation for accurate visual-language alignments.

\subsection{Feature Extraction and Representation\label{subsec:Preliminaries}}

For language representation, we employ a frozen RoBERTa encoder \cite{liu2019roberta}
to extract features $F_{l}\in\mathbb{R}^{L\times C}$, where $L$
denotes the sentence length. For visual presentation, we follow existing
works in the respective tasks. In the RVOS task, we follow \cite{MeViS,heo2022vita,DsHmp}
and utilize the pre-trained Mask2Former architecture \cite{cheng2021mask2former}.
We define $Q_{f}\in\mathbb{R}^{N_{f}\times C}$ as $N_{f}$ learnable
queries to attend to the language feature $F_{l}$ via cross-attention
operation \cite{vaswani2017attention}, producing $\bar{Q_{f}}\in\mathbb{R}^{N_{f}\times C}$.
The Mask2Former processes $\bar{Q_{f}}$ along with $T$ frames of
the input video, generating Mask Features and $\mathcal{K}=\left\{ O_{s}^{t}\right\} _{t=1,s=1}^{T,N_{f}}$
object tokens. These Mask Features will later be applied on top of
the refined object tokens to generate the final segmentation masks
required for the RVOS task. For the VTG task, we use spatio-temporal
features extracted from a pre-trained C3D \cite{tran2015learning}
as the set of object tokens $\mathcal{K}$, following prior works
\cite{2DTAN_2020_AAAI,jiang2019cross,liu2018temporal}.

\subsection{Language-based Step Planner \label{subsec:planner}}

\begin{figure}
\centering{}\includegraphics[width=0.9\columnwidth]{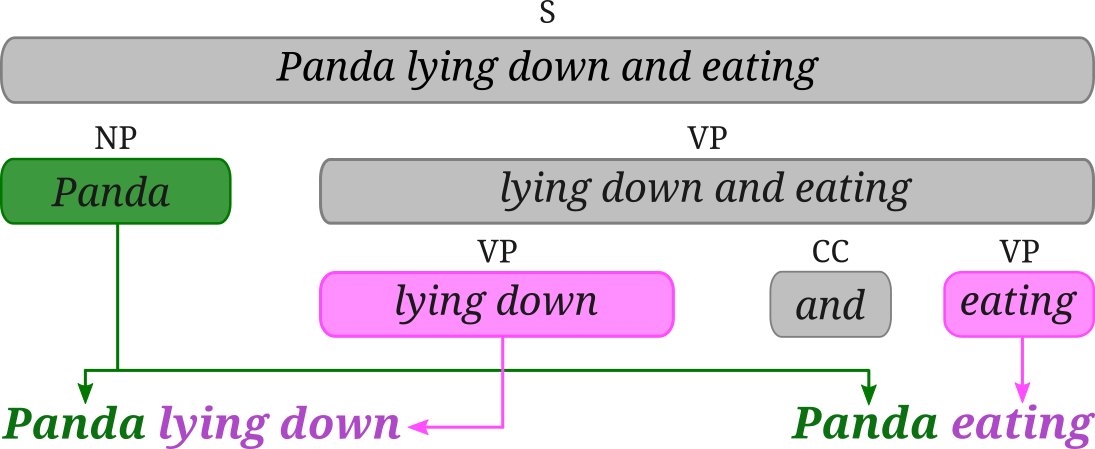}\caption{We use the constituency tree for the decomposition of the sentence
into grammatical components. \label{fig:tree}}
\end{figure}

A fundamental limitation in existing approaches such as \cite{MeViS,heo2022vita,wu2022language}
is the use of single embedding as representations of long sentences.
 While adequate for simple queries, this fails to represent complex
queries describing chains of actions over time. Prior researches \cite{DsHmp,yang2020improving}
show that single embeddings often cannot capture interactions between
multiple objects and entities within complex sentences.

To address this limitation, we introduce Language-based Step Planner,
which decomposes an original prompt of a complex sentence into $P$
sub-prompts, where $P$ adapts to the number of described actions.
The purpose of this decomposition are to break a complex prompt into
simple sub-prompts of simple concepts and actions, enabling more precise
visual-linguistic alignment later at each recurrent refinement step.
For example, as shown in Fig.~\ref{fig:tree}, the sentence \emph{\textquotedblleft Panda
lying down and eating\textquotedblright{}} is decomposed into two
sub-prompts: \emph{\textquotedblleft Panda lying down\textquotedblright{}}
and\emph{ \textquotedblleft Panda eating\textquotedblright }. Each
sub-prompt contains a noun phrase (NP) identifying the target instance
\emph{(Panda)} and a verb phrase (VP) describing its action at the
current step\emph{ (lying down; eating)}.

The decomposition process is detailed in Algorithm \ref{alg:algo_1}.
Given an input sentence, we first generate its constituency parse
tree using the Berkeley Neural Parser \cite{kitaev}. This tree represents
the grammatical structure of the sentence, with nodes labeled based
on their syntactic categories such as noun phrases (NP) and verb phrases
(VP). Through pre-order traversal, we locate the main NP by searching
for the first NP without VP children. This is to ensure the selected
NP represents the central target instance in the input sentence and
not mistakenly picked verb-related modifiers. Once done, the algorithm
continues the pre-order traversal to collect all terminal VP nodes
representing atomic actions. The algorithm then pairs the main NP
with each VP to form complete sub-prompts. Formally, given a sentence
$l$, our Planner generates:
\begin{equation}
L=\text{Planner}(l)=\left\{ \left(N_{p},V_{p}\right)_{p=1}^{P}\right\} ,\label{eq:query}
\end{equation}
where $N_{p},V_{p}$ indicate the NP and VP of the $p^{\text{th}}$
sub-prompt. We encode these NPs and VPs with a pretrained RoBERTa
encoder \cite{liu2019roberta} where the outputted vector representations
of a subject \emph{(e.g., Panda) }varies based on its surrounding
words in each sub-prompt. This encourages our system to capture the
nuanced semantic relationships between the objects in the input sentence.
With slight abuse of notation, we also use $N_{p},V_{p}$ to refer
to the vector representations of the respective NP and VP.

\SetKwComment{Comment}{/* }{ */} 

\begin{algorithm}[h]
\caption{Sentence Decomposition into Sub-prompts of NP-VP Pairs\label{alg:algo_1}}
\label{alg:sentence_decomposition}

\KwIn{ Input sentence $l$. 

}

\KwOut{ Sub-prompts composed of NP-VP pairs: $L=\left\{ \left(N_{p},V_{p}\right)_{p=1}^{P}\right\} $. 

}

\hrulealg

Parse input sentence $l$ into a constituency tree $\mathcal{T}$.\;

$L$$\leftarrow$$\left\{ \right\} $\; 

\tcp{Identify main noun phrase:}

Traverse $\mathcal{T}$ in pre-order, locate the first ``NP'' node\; 

Set $N_{p}$$\leftarrow$ Value of the located NP node.\; 

\tcp{Extract all verb phrases}

Traverse $\mathcal{T}$ in pre-order, locate all \textquotedbl VP\textquotedbl{}
nodes\; 

\ForEach{VP node}{ \lIf{has multiple VP children}{Split them}\lElse{$V_{p}$$\leftarrow$
Value of the located VP node and Add $(N_{p},V_{p})$ pair into $L$}}

\Return $L$\;
\end{algorithm}

\subsection{Language-guided Visual Token Refiner\label{subsec:refiner}}

Our Language-guided Visual Token Refiner module operates as a recurrent
system over $P$ steps, progressively refining the visual token set\emph{
$\mathcal{K}$ }  as guided by the language sequence $L$  prepared
by the Planner. Our key innovation lies in breaking the refinement
process at each recurrent step into two separate stages where spatial
refinement is always followed by temporal refinement. This is inspired
by the English grammar structure in which noun typically comes before
verb in a simple sentence. This saves massive computation compared
to traditional single stage self-attention, while remaining effective
as noted in TimeSformer \cite{bertasius2021space}.

Formally, each recurrent block at stage $p$ updates the representations
of object tokens $\mathcal{O}_{p}$ using two consecutive self-attention
operations as follows:

\begin{align}
\mathcal{\hat{O}}_{p} & =\text{SelfAttnS}\left(\left[\mathcal{O}_{p};\text{Rep}(N_{p})\right]\right),\\
\mathcal{\tilde{O}}_{p} & =\text{SelfAttnT}\left(\left[\mathcal{\hat{O}}_{p};\text{Rep}(V_{p})\right]\right),
\end{align}
where $\left[.\,;.\right]$ denotes the concatenation operation, $\text{Rep}(.)$
replicates the language vector $N_{f}\times T$ times, $\text{SelfAttnS}\left(.\right)$
performs self-attention along the spatial dimension for each video
frame, and $\text{SelfAttnT}\left(.\right)$ performs self-attention
along the temporal dimension for each object. Note that these self-attention
layers operate in different perspectives hence, their weights are
not tied during training. To mitigate the gradient vanishing issue
inherently occurring in recurrent systems, we incorporate a residual
connection \cite{he2016deep} to maintain the information flow:
\begin{equation}
\mathcal{O}_{p+1}\leftarrow W\mathcal{O}_{p}+\mathcal{\tilde{O}}_{p},
\end{equation}
where $W\in R^{C\times C}$ is the learnable weight matrix, and the
initial representation $\mathcal{O}_{0}$ is set to $\mathcal{K}$.

\textbf{Architecture Design Rationale: }This architecture offers three
key advantages. First, disentangling spatial and temporal refinement
enables dedicatedly modeling the interactions of visual tokens of
different kinds through specialized computational units \cite{bertasius2021space}.
Second, it significantly improves computational efficiency when compared
to joint space-time attention. Specifically, each $\text{SelfAttnS}\left(.\right)$
has the complexity of $\ensuremath{O(N_{f}^{2}\times T\times C)}$
and each $\text{SelfAttnT}\left(.\right)$ has the complexity of $\ensuremath{O(N_{f}\times T^{2}\times C)}$,
resulting in the total complexity of $\ensuremath{O(N_{f}^{2}\times T\times C)}+\ensuremath{O(N_{f}\times T^{2}\times C)}=\ensuremath{O(N_{f}T(N_{f}+T)\times C)}$.
This significantly more efficient than the complexity of $\ensuremath{O((N_{f}\times T)^{2}\times C)}$
of the joint space-time attention. Finally, using simple NPs and VPs
to guide these specialized attention layers at each refinement gives
more control over the refinement target, improving the interpretability
of these complex systems.

\subsection{Task-specific Heads and Training Objective\label{subsec:head}}

\begin{figure}
\centering{}\includegraphics[width=0.9\columnwidth]{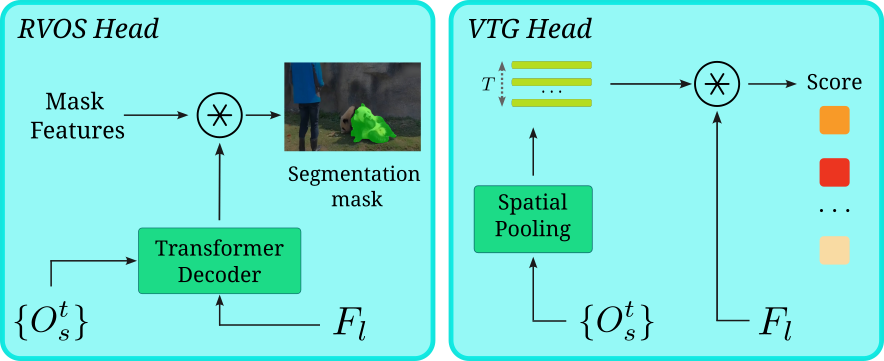}\caption{\textbf{Task-specific heads.} The RVOS head uses language features
$F_{l}$ to attend to refined  visual tokens $\left\{ O_{s}^{t}\right\} $
via a Transformer Decoder, then combines the output with Mask Feature
dot product for final segmentation. For the VTG\emph{ }task,\emph{
}head first applies a spatial pooling operation on each segment and
computes dot product with language features for prediction scores
of correlations between input query and the segments.\label{fig:head}}
\end{figure}

Our framework supports different video-language alignment tasks through
task-specific heads:

\paragraph{Referring Video Object Segmentation (RVOS)}

As shown in Fig.~\ref{fig:head} (left side), the RVOS head uses
a set refined visual tokens $\left\{ O_{s}^{t}\right\} $ via Transformer
Decoder under the language guidance $F_{l}$ and couples with the
Mask features prepared in Sec.~\ref{subsec:Preliminaries} to produce
the final segmentation mask. The combination between the refined visual
tokens and the Mask features is done by the dot product. For the training
objective, we follow \cite{heo2022vita,MeViS} to employ the loss
$\mathcal{L}_{f}$ between the per-frame outputs and the corresponding
frame-wise ground truth, combined with the video-level loss $\mathcal{L}_{v}$
based on video-level ground truth: $\mathcal{L_{\text{RVOS}}}=\lambda_{f}\mathcal{L}_{f}+\lambda_{v}\mathcal{L}_{v}$
where $\lambda_{f}$ and $\lambda_{v}$ are hyper-parameters to numerically
balance the two losses and stabilize the training.

\paragraph{Video Temporal Grounding (VTG)}

As illustrated in Fig.~\ref{fig:head} (right side), the VTG head
begins with a spatial pool operation on all spatial tokens at each
time step to obtain segment representations. These visual features
are then aligned with language features through the dot product to
produce the final prediction scores representing the correlations
between the input query and the segments. The model employs the scaled
$IoU$ as the metric, similar to \cite{2DTAN_2020_AAAI}. For each
segment $s_{i}$, the $IoU$ score $o_{i}$ denotes the overlapping
between $s_{i}$ and the ground truth segment. The final prediction
score $\tilde{y}_{i}$ is obtained by normalizing $o_{i}$ using predefined
thresholds $\tau_{\text{min}}$ and $\tau_{\text{max}}$ as follows:
$\tilde{y}_{i}=\frac{o_{i}-\tau_{\text{min}}}{\tau_{\text{max}}-\tau_{\text{min}}}$,
truncated at $0$ and $1$. We train this task with a binary cross
entropy loss as $\text{\ensuremath{\mathcal{L_{\text{VTG}}}}}=\frac{1}{T}\sum_{i=1}^{T}\left[y_{i}\log\tilde{y}_{i}+(1-y_{i})\log(1-\tilde{y}_{i})\right]$,
where $T$ is the total number of segments in an input video.

\section{$\protect\DataName$ Benchmark}

\begin{table}
\begin{centering}
\begin{tabular}{>{\raggedleft}m{0.34\columnwidth}>{\centering}m{0.1\columnwidth}>{\centering}m{0.1\columnwidth}>{\centering}m{0.1\columnwidth}>{\centering}m{0.1\columnwidth}}
\toprule 
Dataset\hspace*{0.6cm} & Video & Mask & Sentence & ASL\tabularnewline
\midrule
MeViS {\small{}(valid\_u subset)} & 50 & 183 & 793 & 7.7\tabularnewline
$\DataName$\hspace*{0.4cm} & 50 & 183 & 2374 & 17.9\tabularnewline
\bottomrule
\end{tabular}
\par\end{centering}
\caption{Statistics of $\protect\DataName$ and MeViS. $\protect\DataName$,
our newly introduced benchmark, builds upon the videos and mask annotations
from the existing MeViS dataset (valid\_u subset), but replaces the
short language descriptions with complex and descriptive sentences
to evaluate the model's capability in understanding complex compositional
language (ASL: Average Sentence Length).\label{tab:Benchmark}}
\end{table}

Current benchmarks for video-language alignment tasks \cite{seo2020urvos,Gavrilyuk_2018_CVPR}
predominantly focus on short, simple descriptions that fail to validate
models' ability in comprehending complex input sequences. These benchmarks
also often emphasize static appearance factors over temporal dynamics,
thus inadequate to test object evolution in videos. While MeViS \cite{MeViS}
introduced motion-focused video descriptions, its language annotations
remain limited to simple expressions typically describing only one
or two actions.

To address these limitations, we introduce $\DataName$ (which stands
for \textbf{\uline{MeViS}} with e\textbf{\uline{X}}tended compositional
queries), a benchmark designed to target the evaluation of vision-language
alignments in videos on complex, extended queries. We build upon the
video content and mask annotations from MeViS's valid\_u subset but
extend the original simple descriptions into fully descriptive and
complex compositional sentences. Following the existing practice \cite{trackgpt,MeViS,bai2024one},
we employ a LLM to paraphrase the language expressions for augmentation,
followed by manual verification to ensure correctness and diversity.
Through this process, we establish a comprehensive benchmark assess
model performance in handling complex queries. As detailed in Table~\ref{tab:Benchmark},
$\DataName$ includes $2374$ query descriptions, with an Average
Sentence Length (ASL) of 17.9 words, significantly exceeds the 7.7
words of the MeViS dataset.

To ensure data quality, we follow the procedure outlined in \cite{le2024progressive},
randomly sampling $2\%$ of the data for evaluation by three independent
human annotators. The evaluators assessed three key aspects of the
textual descriptions: naturalness (\textquotedbl Does the generated
description sound natural?\textquotedbl ), ambiguity (\textquotedbl Is
the sentence clear and unambiguous?\textquotedbl ), and correctness
of the segmentation masks (\textquotedbl Is the segmentation mask
correct for the query?\textquotedbl ). The assessment results show
that $85.7\%$ of the text descriptions were rated as natural, $98.9\%$
accurately described the segmentation masks, and $95.2\%$ of generated
sentences were clear. We measured inter-annotator agreement using
Cohen's Kappa scores \cite{cohen1960coefficient}, achieving an average
score of $0.8053$ for naturalness, $0.7724$ for accuracy, and $0.8607$
for clarity. These scores indicate a high level of agreement among
annotators.

\section{Experiments\label{sec:Experiments}}

\subsection{Referring Video Object Segmentation}

\paragraph{Dataset}

We evaluate our method on MeViS \cite{MeViS}, which particularly
focuses on perceiving object motion. The dataset contains 2,006 videos
and 28,570 sentences, divided into train, valid\_u, validation, and
test. Following standard practice \cite{MeViS,DsHmp}, we report performance
on the validation subset, while using the valid\_u for qualitative
analysis.

\paragraph{Evaluation metrics}

Following prior works \cite{MeViS,seo2020urvos,ding2022language,wu2022language,DsHmp},
we employ two primary metrics: region similarity ($\mathcal{J}$)
and contour accuracy ($\mathcal{F}$). $\mathcal{J}$ measures the
Intersection over Union $(IoU)$ between predicted and ground truth
masks, while $\mathcal{F}$ evaluates contour accuracy of the prediction.
The main evaluation metric $\mathcal{J\text{\&}F}$ is the average
of these two.

\paragraph{Training details}

We adopt the training configuration from LMPM \cite{MeViS}. Specifically,
we train our network for $50,000$ iterations on four A100 40GB GPUs
with AdamW optimizer \cite{loshchilov2017decoupled} with a initial
learning rate of $1\times10^{-5}$. We sample number of frame $T=6$
as set $N_{f}=20$. We use the Swin-Tiny backbone for the Mask2Former
vision encoder. For the text encoder, we use a frozen RoBERTa \cite{liu2019roberta}
model. During training and evaluation, all frames are cropped so that
the longest side measures $640$ pixels and the shortest side measures
$360$ pixels. 

\subsubsection{Quantitative results on $\protect\DataName$ benchmark \label{subsec:long_query}}

\begin{table}
\begin{centering}
{\footnotesize{}}%
\begin{tabular}{>{\raggedright}m{0.3\columnwidth}>{\raggedright}m{0.2\columnwidth}>{\centering}m{0.09\columnwidth}>{\centering}m{0.09\columnwidth}>{\centering}m{0.09\columnwidth}}
\toprule 
Method\hspace*{0.6cm} & Reference & $\mathcal{J\text{\&}F}$ & $\mathcal{J}$ & $\mathcal{F}$\tabularnewline
\midrule
\midrule 
LMPM \cite{MeViS} & ICCV 23 & 39.9 & 37.2 & 42.6\tabularnewline
DsHmp \cite{DsHmp} & CVPR 24 & 50.6 & 48.2 & 53.0\tabularnewline
HTR \cite{miao2024htr} & TCSVT 24 & 44.7 & 48.3 & 41.1\tabularnewline
VideoLISA \cite{bai2024one} & NeurlPS 24 & 46.9 & 47.2 & 46.6\tabularnewline
\midrule 
$\ModelName$ & \_ & 53.8 & 52.9 & 54.7\tabularnewline
\bottomrule
\end{tabular}{\footnotesize\par}
\par\end{centering}
\caption{Performance comparison on $\protect\DataName$ benchmark \label{tab:LongQuery}}
\end{table}

\begin{figure}[t]
\includegraphics[width=0.9\columnwidth]{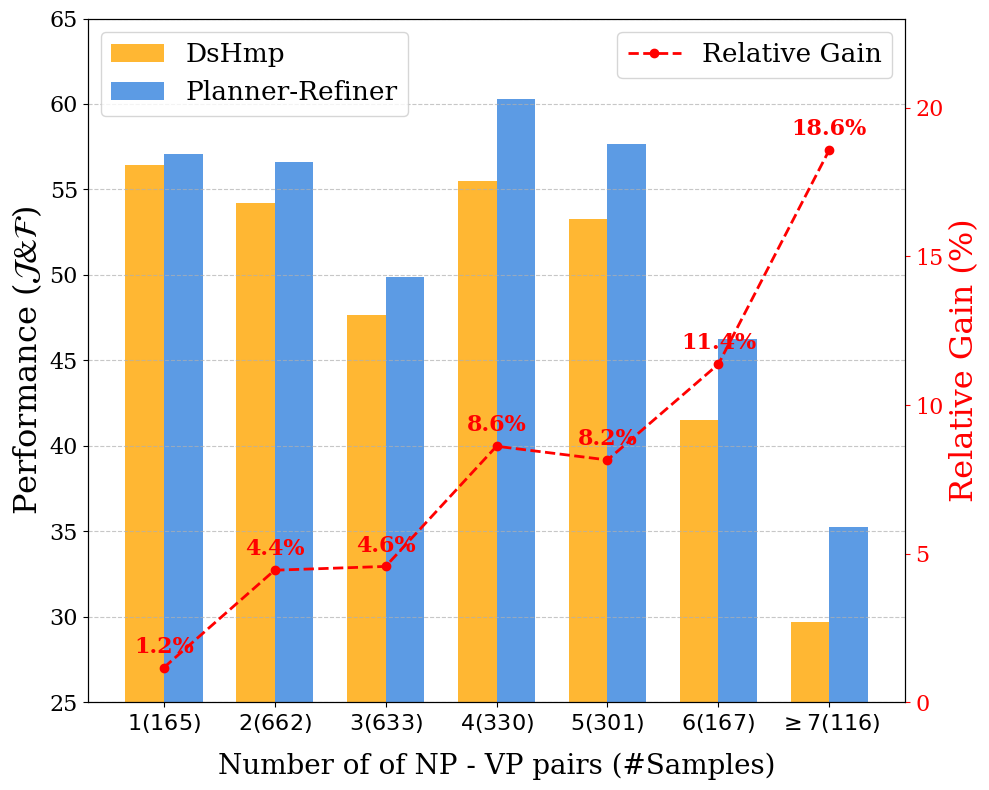}\caption{Performance across sentence components on the $\protect\DataName$
benchmark. The x-axis shows the number of components (NP-VP pairs)
with the corresponding sample count in parentheses. Performance is
compared between DsHmp and $\protect\ModelName$ using bar charts,
while the relative improvement of $\protect\ModelName$ over DsHmp
is depicted by the red line. \label{fig:ablation_2}}
\end{figure}

To evaluate performance on complex queries, we compare $\ModelName$
against state-of-the-art RVOS methods on the introduced $\DataName$
benchmark. All methods are trained on MeViS and evaluated on $\DataName$
in a zero-shot setting. We report the results in Tab.~\ref{tab:LongQuery}.
LMPM, which processes entire sentences at once, achieves less than
$40$ $\mathcal{J\text{\&}F}$. DsHmp advances the prior work by decoupling
static (noun) and motion (verb) components, but the fixed two-part
decomposition limits its capacity to handle linguistically complex
inputs. HTR is a recent method that introduces memory components into
visual processing. However, it still fails to effectively address
complex queries, achieving a performance of just under $45$ $\mathcal{J\text{\&}F}$
on $\DataName$. More recently, VideoLISA pioneers the use of large
language models for language-instructed video segmentation. While
incorporating a large language model enhances reasoning capability,
the absence of an effective mechanism for handling complex queries
results in weak performance on the $\DataName$ benchmark. In contrast,
our method, equipped with a novel recursive refinement mechanism,
achieves $53.8$ $\mathcal{J\text{\&}F}$ ---significantly outperforming
all prior approaches.

Fig.~\ref{fig:ablation_2} provides detailed analysis across query
lengths for DsHmp and our $\ModelName$. Both methods show declining
performance with increased sentence complexity, with performance of
below $40$$\mathcal{J\text{\&}F}$ for queries containing more than
$6$ NP-VP pairs in comparison to $50\lyxmathsym{\textendash}60$$\mathcal{J\text{\&}F}$
for sentences having $1$ to $4$ components. 

As we can see, both methods $\ModelName$ and DsHmp show comparable
performance with short queries. However, $\ModelName$ achieves clearly
increasing advantages with query complexity - nearly $10\%$ in the
case of 4 components and rising to nearly $20\%$ for those with over
$6$ components. These results clearly highlight the effectiveness
of $\ModelName$ in handling complex, compositional descriptions in
vision-language grounding tasks in videos.

\subsubsection{Quantitative results on the MeViS validation set}

While exhibiting superior performance on the long query dataset, our
method maintain competitive performance on standard RVOS dataset.
As shown in Tab.~\ref{tab:quantitative}, our method $\ModelName$
consistently outperforms existing approaches on MeViS. Early methods
like VLT+TC \cite{9932025} and URVOS \cite{seo2020urvos} adapted
image-based techniques to videos through frame-by-frame segmentation
and temporal post-processing, but failed to effectively capture motion
dynamics. While recent approaches including DETR-based structures
such as MTTR \cite{botach2022end}, LBDT \cite{ding2022language}
showed improvements, they still lack explicit motion modeling.

The baseline LMPM \cite{MeViS} introduced more advanced temporal
modeling, but the reliance on single-stage self-attention limits its
effectiveness. Our approach delivers a significant $8.8$ $\mathcal{J\text{\&}F}$
improvement (over $20$\% relative gain) through the recurrent mechanism.
Compared to state-of-the-art DsHmp \cite{DsHmp}, ours achieves a
slightly better overall performance ($46.0$ vs. $45.1$). However,
$\ModelName$ demonstrates particular advantages in handling complex
queries (Refer to Tab.~\ref{tab:LongQuery} and Fig.~\ref{fig:ablation_2}).
To ensure a fair comparison, all methods are trained and evaluated
using identical vision and language encoders, along with the same
loss function $\mathcal{L_{\text{RVOS}}}$ as defined in Sec.~\ref{subsec:head}.

We also compare $\ModelName$ against recent studies including HTR
\cite{miao2024htr}, which employs memory components, and VideoLISA
\cite{bai2024one}, which leverages LLMs for enhanced reasoning capabilities.
The results show that these models clearly struggle to handle complex
descriptions of real-world events involving multiple fine-grained
actions while ours addresses these more effectively.

\begin{table}
\begin{centering}
{\footnotesize{}}%
\begin{tabular}{>{\raggedright}m{0.28\columnwidth}>{\raggedright}m{0.16\columnwidth}>{\centering}m{0.1\columnwidth}>{\centering}m{0.1\columnwidth}>{\centering}m{0.1\columnwidth}}
\toprule 
Method\hspace*{0.6cm} & Reference & $\mathcal{J\text{\&}F}$ & $\mathcal{J}$ & $\mathcal{F}$\tabularnewline
\midrule
\midrule 
URVOS \cite{seo2020urvos}\hspace*{0.4cm} & ECCV 20 & 27.8 & 25.7 & 29.9\tabularnewline
LBDT \cite{ding2022language}\hspace*{0.4cm} & CVPR 22 & 29.3 & 27.8 & 30.8\tabularnewline
MTTR \cite{botach2022end}\hspace*{0.4cm} & CVPR 22 & 30.0 & 38.8 & 31.2\tabularnewline
ReferFormer \cite{wu2022language}\hspace*{0.4cm} & CVPR 22 & 31.0 & 39.8 & 32.2\tabularnewline
VLT+TC \cite{9932025}\hspace*{0.4cm} & TPAMI 23 & 35.5 & 33.6 & 37.3\tabularnewline
LMPM \cite{MeViS}\hspace*{0.4cm} & ICCV 23 & 37.2 & 34.2 & 40.2\tabularnewline
DsHmp \cite{DsHmp}\hspace*{0.4cm} & CVPR 24 & 45.1 & 41.8 & 48.4\tabularnewline
HTR \cite{miao2024htr} \hspace*{0.4cm} & TCSVT 24 & 42.7 & 39.9 & 45.5\tabularnewline
VideoLISA \cite{bai2024one} \hspace*{0.4cm} & NeurlPS 24 & 44.4 & 41.3 & 47.6\tabularnewline
\midrule 
$\ModelName$ & \_ & 46.0 & 44.5 & 47.5\tabularnewline
\bottomrule
\end{tabular}{\footnotesize\par}
\par\end{centering}
\caption{Performance comparison on MeViS dataset.\label{tab:quantitative}}
\end{table}

\subsubsection{Qualitative Results}

\begin{figure*}
\begin{centering}
\includegraphics[width=0.85\textwidth]{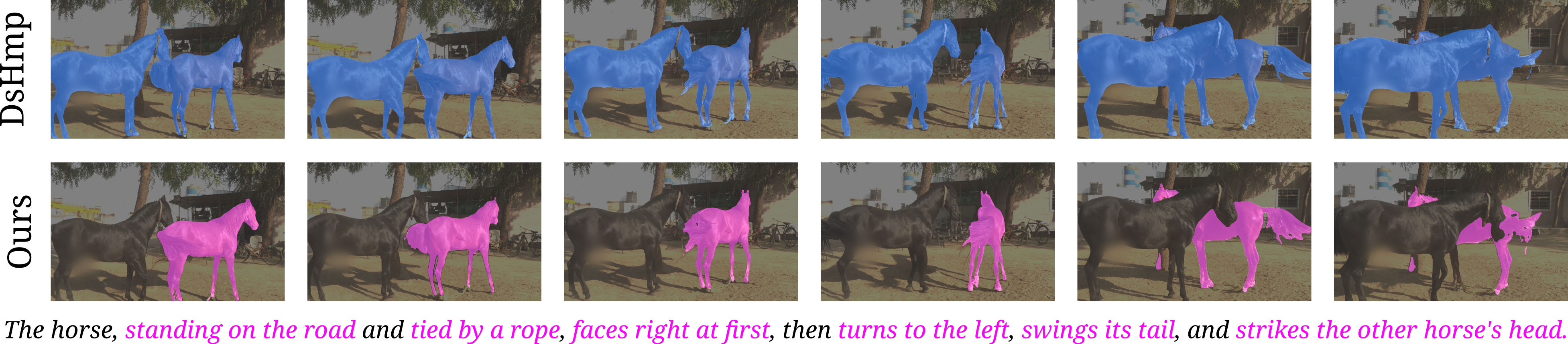}
\par\end{centering}
\caption{Qualitative Comparison: DsHmp (Blue masks) and Ours (Pink masks);
the pink masks correctly match the ground truth.\label{fig:Qualitative-results}}
\end{figure*}

Fig.~ \ref{fig:Qualitative-results} presents qualitative comparisons
between our method and DsHmp on a representative example from the
MeViS-X benchmark. The result demonstrate that our method effectively
captures the motion-related component (highlighted in pink in the
sentence), enabling accurate identification of the correct horse described
by the complex linguistic description. In contrast, DsHmp almost relies
only on static cues \emph{(Horse)} from language to localize the target
object, demonstrating limited ability to comprehend motion-related
information (e.g., \emph{\textquotedbl turn to the left,\textquotedbl{}
\textquotedbl swing tail,\textquotedbl{} \textquotedbl strike\textquotedbl ,}
$\dots$).

\begin{figure*}
\begin{centering}
\includegraphics[width=0.85\textwidth]{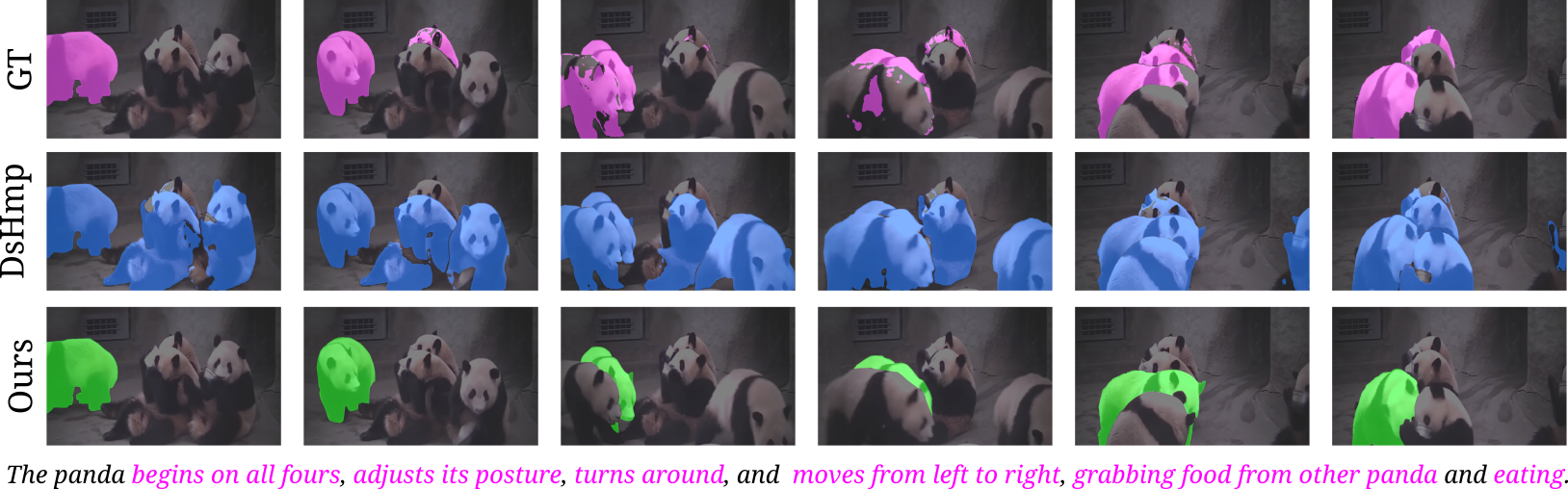}
\par\end{centering}
\caption{Error Case Analysis: Green, Blue, and Pink masks denote results of
Ground Truth, DsHmp, and our method, respectively. \label{fig:Fail_mode}}
\end{figure*}

\subsubsection{Error Case Analysis}

To better understand the model's behavior, we analyze an error case
and present the results in Fig. \ref{fig:Fail_mode}. This is a challenging
scenario due to the presence of multiple potential target objects\emph{
(Panda)}, which are closely located and overlap with each other. Both
methods produce false positives by including parts of other pandas
in the segmentation masks. While our method successfully identifies
the correct panda, it occasionally segments parts of surrounding pandas
that exhibit similar behaviors (e.g., \emph{moving, eating}). This
observation highlights the need for a more robust approach to accurately
pinpoint the correct object in crowded scenarios.

\subsubsection{Ablation Studies}

\begin{table}
\begin{centering}
\begin{tabular}{>{\centering}m{0.09\columnwidth}>{\raggedleft}m{0.32\columnwidth}>{\centering}m{0.1\columnwidth}>{\centering}m{0.1\columnwidth}>{\centering}m{0.1\columnwidth}}
\toprule 
Index & Method\hspace*{0.6cm} & $\mathcal{J\text{\&}F}$ & $\mathcal{J}$ & $\mathcal{F}$\tabularnewline
\midrule
\midrule 
0 & Baseline & 37.2 & 34.2 & 40.2\tabularnewline
1 & W/o $\text{SelfAttnS}\left(\right)$ & 41.8 & 40.0 & 43.6\tabularnewline
2 & W/o $\text{SelfAttnT}\left(\right)$ & 40.8 & 38.1 & 43.5\tabularnewline
3 & W/o Lang. Concat. & 39.8 & 37.4 & 42.2\tabularnewline
4 & Joint ST Attention & 42.0 & 39.2 & 44.8\tabularnewline
5 & Swapping VP and NP & 44.6 & 43.7 & 45.5\tabularnewline
6 & Averaging & 44.7 & 42.5 & 46.9\tabularnewline
7 & Sum & 43.8 & 41.6 & 46.0\tabularnewline
\midrule 
 & $\ModelName$ & 46.0 & 44.5 & 47.5\tabularnewline
\bottomrule
\end{tabular}
\par\end{centering}
\caption{Ablation studies on the validation set of MeVis dataset\label{tab:Ablation}}
\end{table}

We performed ablation studies to gain insights into our recurrent
system (See Tab. \ref{tab:Ablation}). In particular, we ablate or
modify individual components to evaluate their impact on the overall
performance. All experiments use the Swin-Tiny backbone and evaluate
on the validation set.

\emph{Effectiveness of Spatial Refinement:} Removing the self-attention
along the spatial dimension ($\text{SelfAttnS}\left(.\right)$), leading
to a performance drop of over $4.0$ points, underscoring the critical
role of spatial refinement.

\emph{Effectiveness of Temporal Refinement:} Similarly, removing the
temporal refinement step ($\text{SelfAttnT}\left(.\right)$), resulting
in a performance drop of over $5.0$ points. This decline highlights
the necessity of this component for the model's temporal understanding
capability.

\emph{Effectiveness of Language Guidance:} Removing language features
NP and VP during refinement causes performance to degrade to below
$40$ $\mathcal{J\text{\&}F}$, highlighting the crucial role of language
information in the refinement process.

\emph{Effectiveness of Two-Stage Attention:} We observe a performance
drop of $4.0$ points when doing simultaneous refinement along spatial
and temporal dimensions, which also aligns with the findings in TimeSformer.
\cite{bertasius2021space}.

\emph{Effect of Swapping VP and NP Order: }Placing the VP before the
NP results in a noticeable drop in $\mathcal{J\text{\&}F}$ score
(from $46.0$ to $44.6$), supporting our intuition that spatial refinement
should precede temporal refinement. This is because spatial modeling
establishes object existence, while temporal modeling captures the
evolution of object dynamics over time.

\emph{Effectiveness of Recurrent Design:} To evaluate the importance
of recurrent processing, we test an alternative where sub-prompts
independently refine visual tokens, with outputs combined through
summation or averaging, rather than our progressive refinement approach.
This parallel processing strategy decreases performance by $1.0$
-- $2.0$ points, demonstrating the effectiveness of our recurrent
mechanism.

\subsection{Temporal Grounding}

\paragraph*{Dataset}

We use the TACoS dataset to evaluate the proposed method on the Temporal
Grounding application. This dataset includes 127 videos taken from
the MPII Cooking Composite Activities video corpus \cite{rohrbach2012script},
showcasing various activities in a kitchen setting. A standard split
\cite{gao2017tall} comprises 9,790 samples for training, 4,436 for
validation, and 4,001 for testing.

\paragraph*{Evaluation Metrics}

Following previous works \cite{2DTAN_2020_AAAI,gao2017tall,chen2019semantic,jiang2019cross,yuan2019find},
we use $Rank\,n\text{@}m$, which measures the percentage of queries
where correct predictions appear in top-$n$ results. A prediction
is correct if its temporal segment overlaps with the ground truth
exceed $m$ (in per cent). We report results for $n\in\{1,5\}$ and
$m\in\{0.1,0.3,0.5\}$. 

\paragraph*{Training}

We train our model with the Adam optimizer \cite{loshchilov2017decoupled}
with a learning rate of $10^{-4}$ and a batch size of 64. For a fair
comparison, we utilize the same visual features, which is C3D \cite{tran2015learning},
as previous studies. The total number of sampled clips $T=128$. Each
clip consists of $16$ frames, with an overlap of $0.8$ between consecutive
clips and having spatial tokens $N_{f}=512$. The scaling thresholds
$\tau_{\text{min}}$ and $\tau_{\text{max}}$ are set to $0.3$ and
$0.7$, respectively.

\subsubsection{Quantitative Results}

\begin{table}
\begin{centering}
\begin{tabular}{>{\centering}m{0.07\textwidth}>{\centering}m{0.09\columnwidth}>{\centering}m{0.09\columnwidth}>{\centering}m{0.09\columnwidth}>{\centering}m{0.09\columnwidth}>{\centering}m{0.09\columnwidth}>{\centering}m{0.09\columnwidth}}
\toprule 
\multirow{2}{0.07\textwidth}{Method} & \multicolumn{3}{c}{$\textit{Rank}1\text{@}$} & \multicolumn{3}{c}{$\textit{Rank}5\text{@}$}\tabularnewline
\cmidrule{2-7} \cmidrule{3-7} \cmidrule{4-7} \cmidrule{5-7} \cmidrule{6-7} \cmidrule{7-7} 
 & 0.1 & 0.3 & 0.5 & 0.1 & 0.3 & 0.5\tabularnewline
\midrule
SAP  & 31.2 & \textminus{} & 18.2 & 53.5 & \textminus{} & 28.1\tabularnewline
SLTA  & 23.1 & 17.1 & 11.9 & 46.5 & 32.9 & 20.9\tabularnewline
ABLR  & 34 .7 & 19.5 & 9.4 & \textminus{} & \textminus{} & -\tabularnewline
CBLN & 49.1 & 38.9 & 27.6 & 73.1 & 59.9 & 46.2\tabularnewline
2D-TAN  & 47.6 & 37.3 & 25.3 & 70.3 & 57.8 & 45.0\tabularnewline
\midrule 
Ours & 51.2 & 40.6 & 28.7 & 78.6 & 62.6 & 49.8\tabularnewline
\bottomrule
\end{tabular}
\par\end{centering}
\caption{Performance comparison on TACoS dataset}
\end{table}

We evaluate the performance of the proposed $\ModelName$ model against
existing methods, including SAP \cite{chen2019semantic}, SLTA \cite{jiang2019cross},
ABLR \cite{yuan2019find}, CBLN \cite{liu2021context} and 2D-TAN
\cite{2DTAN_2020_AAAI}. While these methods adopt various strategies
to align visual and linguistic domains, they lack a robust mechanism
to effectively address complex, multi-component sentence descriptions.
In contrast, our approach introduces a novel solution by decomposing
the original sentence into multiple sub-prompts to refine the visual
token set in a recursive manner. The consistent improvements over
all evaluation metrics underscore the potential of the proposed approach.

\section{Conclusion}

In this work, we propose a novel method for grounding visual information
in videos based on referring descriptions. Unlike previous approaches
that rely on single-stage refinement with simplified sentence embeddings,
our method decomposes the input sentence into sub-prompts composed
of noun-phrase and verb-phrase pairs, and progressively refines visual
tokens through language-guided recurrence. Extensive quantitative
and qualitative evaluations on two different video grounding tasks
demonstrate the effectiveness of our approach. Additionally, we introduce
a new benchmark, $\DataName$, designed to assess model performance
under extended descriptions, further validating the effectiveness
of our method across diverse alignment scenarios.

\bibliography{mybibfile}

\end{document}